\pdfoutput=1
\documentclass[11pt]{article}
\usepackage{url}
\usepackage[final]{acl}

\usepackage{savesym}
\savesymbol{dddot}
\savesymbol{ddddot}

\usepackage[utf8]{inputenc}
\usepackage[T1]{fontenc}
\usepackage{times}
\usepackage{latexsym}
\usepackage{kotex}
\usepackage{amsmath}
\usepackage{amsfonts}
\usepackage{array}
\usepackage{booktabs}
\usepackage{epsfig}
\usepackage{graphicx}
\usepackage{longtable}
\usepackage{makecell}
\usepackage{multirow}
\usepackage{pgfplots}[style=base]
\pgfplotsset{compat=1.18}
\usepackage{soul}
\usepackage{subcaption}
\usepackage{tabularx}
\usepackage{tabu}
\usepackage{threeparttable}
\usepackage{xcolor,colortbl}
\usepackage{bbold}
\usepackage{xspace}
\usepackage{supertabular}
\usepackage{multicol}
\usepackage{minitoc}
\usepackage{tcolorbox}
\usepackage{microtype}
\usepackage{inconsolata}
\usepackage{accents}
\usepackage{changepage}   
\usetikzlibrary{fit,tikzmark,positioning}

\title{Decoding the Rule Book: Extracting Hidden Moderation Criteria from Reddit Communities \thanks{Accepted to EMNLP 2025 Main.}\\
\small \textcolor{red}{WARNING: The content contains model outputs that are offensive and toxic.}
}

\author{
  \textbf{Youngwoo Kim\textsuperscript{1}}, \textbf{Himanshu Beniwal\textsuperscript{1, 2}},
  \textbf{Steven L. Johnson\textsuperscript{1}},
  \textbf{Thomas Hartvigsen\textsuperscript{1}}
  \\
  \textsuperscript{1}University of Virginia,
  \textsuperscript{2}Indian Institute of Technology Gandhinagar
  \\
  \small{
  \texttt{youngwookim@cs.umass.edu, himanshubeniwal@iitgn.ac.in,}} \\
  \small{\texttt{steven@virginia.edu, hartvigsen@virginia.edu}}
}

\begin{document}
\maketitle

\begin{abstract}
Effective content moderation systems require explicit classification criteria, yet online communities like subreddits often operate with diverse, implicit standards. This work introduces a novel approach to identify and extract these implicit criteria from historical moderation data using an interpretable architecture. 
We represent moderation criteria as score tables of lexical expressions associated with content removal, enabling systematic comparison across different communities.
Our experiments demonstrate that these extracted lexical patterns effectively replicate the performance of neural moderation models while providing transparent insights into decision-making processes. The resulting criteria matrix reveals significant variations in how seemingly shared norms are actually enforced, uncovering previously undocumented moderation patterns including community-specific tolerances for language, features for topical restrictions, and underlying subcategories of the toxic speech classification. 
\end{abstract}

\newcommand{\mytable}{CriteriaMatrix\xspace}

\newcommand{\todo}[1]{{\color{red} (TODO: #1) \\}}
\section{Introduction}
% (Data background)
Content moderation is essential for fostering healthy online discourse, but remains challenging due to the diverse and often implicit norms that govern different communities.
While platforms like Reddit\footnote{\url{https://www.reddit.com/}} host numerous micro-communities (subreddits) with distinct norms and moderation practices, the specific criteria used to enforce these norms often remain opaque.

Previous research has examined emergent norms across Reddit communities, identifying both generic norms (shared across many communities) and community-specific norms (\autoref{tab:rules}) ~\cite{fiesler2018reddit, chandrasekharan2018internet, park2024valuescope}. However, even when communities share similar norm categories (e.g., ``Be civil''), the precise criteria for violations differ between communities based on moderator decisions, topic domains, and user expectations. These criteria are rarely fully captured in stated rules alone, instead requiring substantial domain knowledge and familiarity with community practices.

While existing approaches have leveraged historical moderation data to train classifiers that predict norm violations~\cite{chandrasekharan2019crossmod, park2021detecting}, these models often function as black boxes, making it difficult to understand the specific patterns being used for classification decisions. This opacity presents significant challenges for practical implementation, as moderation systems must reflect moderators' intent~\cite{kolla2024llm} rather than merely detecting superficial features.

% Content moderation aims to replicate moderators' intent, with classification criteria that are highly dependent on broader context and often not explicitly stated. Thus, external annotators may struggle to predict the ideal labels that community moderators would assign.
% Such difficulty amplify when one blindly train blackbox classifiers for moderation predictions, as features used by the model may be superficial and not adaptable to moderators' expectations.

\begin{figure*}[t]
    \centering
    \includegraphics[width=\linewidth]{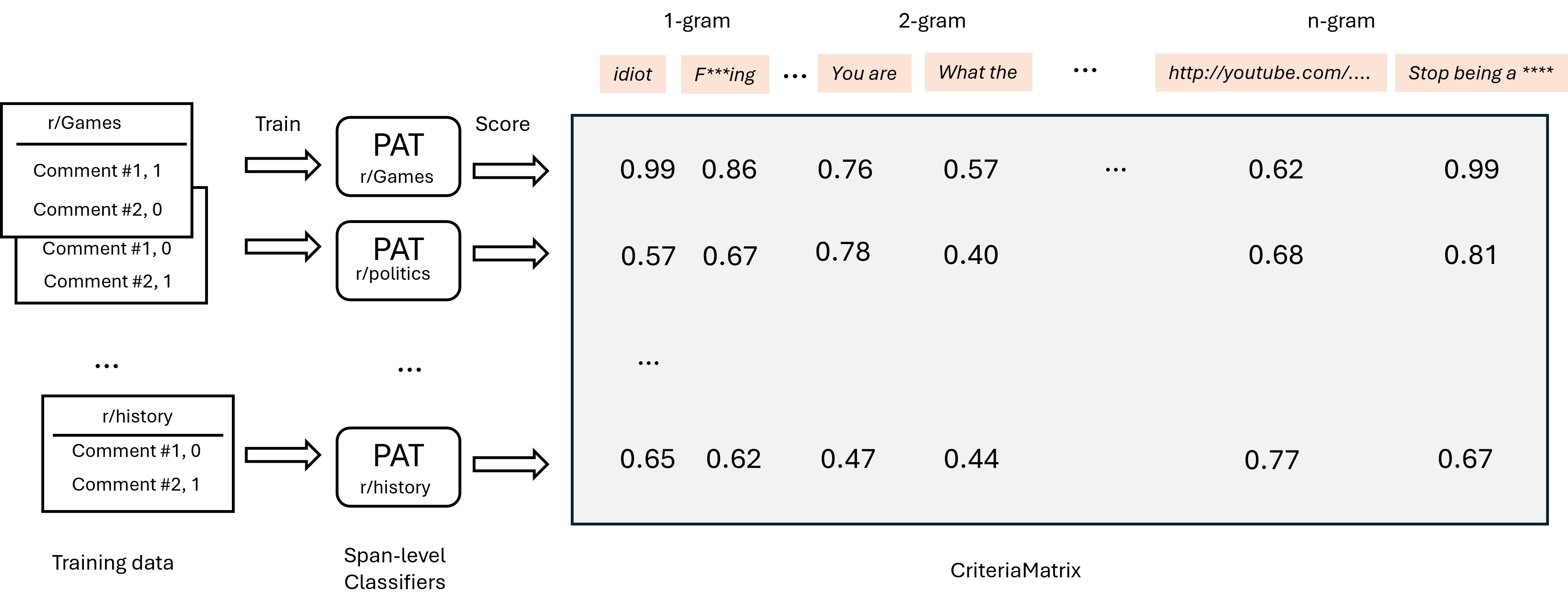}
    \caption{Overview of criteria discovery. For each subreddit, a text classifier (PAT) is trained from past moderation data of comment deletion. PAT models for each subreddit score each term in the vocabulary, allowing us to compare across different subreddits.}
    \label{fig:outline}
\end{figure*}

In this work, we aim to explicitly discover the implicit criteria used in moderation decisions across Reddit communities. We introduce an approach that extracts and represents these criteria as score tables of lexical expressions associated with content removal. We propose to build this \mytable by employing Partial Attention Transformer (PAT)~\cite{condnli}, an interpretable architecture that can predict moderation probabilities on any lexical expressions. . 
For each of individual communities, a PAT model is trained with corresponding data, and used to predict moderation scores for a given lexical expression in the corresponding community. This enables comparison of moderation patterns across communities.

Our experiments on the Reddit moderation dataset~\cite{chandrasekharan2018internet} show that interpretable model PAT can effectively replicate the performance of neural moderation models, achieving comparable results to ChatGPT. PAT is further used to build analysis using PAT provides various insights of implicit criteria that drive moderation decisions, providing hints of potential risks and future directions for better moderation system.

Our contributions include: (1) a novel representation of moderation criteria as scores assigned to lexical expressions, providing unambiguous and verifiable insights; (2) demonstrating how PAT can effectively identify global moderation patterns by assigning well-calibrated scores to text spans; and (3) revealing previously unrecognized characteristics of moderation practices across Reddit communities, including varying tolerances for similar content types and community-specific enforcement patterns.

\begin{table}[h]
\small
\centering
\begin{tabular}{p{3.5cm} p{3cm}}
\toprule
\textbf{Violation Category} & \textbf{Example Rule} \\
\midrule
Incivility & \textit{“Be civil”} \\
Harassment, Doxxing & \textit{“Don’t harass others”} \\
Spam, Reposting, Copyright & \textit{“No excessive posting”} \\
Format, Images, Links & \textit{“Use the correct tags”} \\
Low-quality, Spoilers & \textit{“No low-quality posts”} \\
Off-topic, Politics & \textit{“Only relevant posts”} \\
Hate Speech & \textit{“No racism, sexism”} \\
Trolling, Personal Army & \textit{“No trolls or bots”} \\
Meta-Voting & \textit{“No downvoting”} \\
\bottomrule
\end{tabular}
\caption{Common subreddit norm violations and paraphrased rules \cite{park2021detecting}}
\label{tab:rules}
\end{table}

% Backgorund
% There were efforts to understand the norms that abstract the moderation rules and patterns~\cite{chandrasekharan2018internet, park2021detecting, fiesler2018reddit}.
% They discovered the shared norms such as prohibiting hate speeches, and specific norms that limit which topics are acceptable.
% For shared norms, there were efforts to curate the dataset and develop classifiers either in the context of micro-communities~\cite{park2021detecting} or for larger platforms~\cite{inan2023llama, singhal2023sok}.

 %(Addressing limitations of naïve approaches)
% Here is an example of a moderation criterion we identified. While it is known that blaming moderators often leads to content removal, the specific threshold for what constitutes \emph{blaming} has been unclear. Our method reveals that simply mentioning the word \texttt{mods}---a common reference to volunteer moderators---is a strong predictor of moderation.

% Our system aims to characterize the differences in moderation policies across subreddits, providing insights that can help future developers build more controllable and trustworthy AI moderation systems.

\section{Related works}

\subsection{Criteria extraction for content moderation.}

Research on online community standards has identified underlying norms that guide content moderation across platforms~\cite{chandrasekharan2018internet, park2024valuescope, neuman2023ai}. These norms typically represent high-level concepts, while community rules provide more explicit, fine-grained enforcement guidance~\cite{park2021detecting, rao-etal-2023-makes}.

Prior works on norm discovery exhibit several notable limitations.
First, many approaches are constrained by limited categories of norms, such as (e.g., supportiveness, politeness, humor)~\cite{park2024valuescope,goyal2024uncovering}, which restricts the discovery of novel norm types.

Second, both norms and rules remain abstract, with the specific criteria used in violation decisions often remaining vague or implicit. Previous approaches to determine whether a given text violates a particular norm have relied on assessments from crowd-workers or generic LLMs~\cite{neuman2023ai, park2024valuescope}. However, these assessment methods likely deviate from the actual criteria applied by moderators.

Lastly, the extracted norms are not evaluated if they are actually predictive of moderation outcomes~\cite{chandrasekharan2018internet, neuman2023ai}, while only parts of extraction pipeline is evaluated.

Interviewing or surveying on actual moderators and users are informative~\cite{lambert2024positive, weld2024making}, but they focused on commonality of aspects and less on identifying fine-grained criteria differences.

\subsection{Model interpretability}
Our work approaches criteria extraction as a feature extraction problem. We train a model and interpret it to gain insights about the data, which is relevant to the broader field of model interpretability. Specifically, we choose PAT~\cite{condnli}, which is an interpretable architecture and demonstrated to be capable of discovering dataset or model biases in information retrieval domain~\cite{kim2024discovering}. There are a few reasons that we chose PAT over alternatives. While most interpretability methods like LIME~\cite{lime} and SHAP~\cite{lundberg2017unified} focus on local explanations for specific instances~\cite{burkart2021survey}, we require global explanations that characterize the entire model~\cite{phillips2018interpretable, kim2024discovering}. Local explanation methods have significant drawbacks for our purposes: they are computationally expensive~\cite{ahmed2024comparative}, require multiple forward passes per instance, and produce scores that are not meaningful as a probability. 

\subsection{Content moderation systems}
Large portion of works in content moderation are focused on incivility (toxicity) and hate speech~\cite{park2021detecting}, which are not sufficient to capture diverse norms in micro-communities like Reddit.  A notable approach for capturing diverse norms involves using moderators' rule violation comments as training signals, allowing models to learn from explicit moderation decisions~\cite{park2021detecting}. However, it is limited to few communities where explanation comments are provided. Moreover, it is not clear if a violation criteria for a norm in one community is applicable to others. 

Recent work has explored using instruction-following LLMs like ChatGPT for moderation rule enforcement~\cite{kolla2024llm}, but our experiments confirm previously reported limitations: these models achieve only moderate accuracy and struggle with community-specific rules beyond universal violations like incivility and hate speech.

 There were concern that content moderation models with seemingly high performance are indeed relying on the mentions of the ethnic group and spurious features~\cite{rottger2021hatecheck, hartvigsen2022toxigen}. Our work address this issue by discovering actual criteria used for classifications, which allow us to identify spurious features or potential limitations of models.
 
\section{Criteria discovery}
\label{sec:pat}

\subsection{Overview}
%< Subreddit specific analysis step >
We conceptualize the challenge of understanding community-specific moderation as a vocabulary scoring problem.
While moderation decisions are complex and contextual, we hypothesize that they can be meaningfully represented through predictive lexical patterns that signal rule violations. Our approach aims to build explicit, unambiguous representations of subreddit-specific moderation criteria by extracting and scoring phrasal expressions for each communities. This vocabulary-based representation offers several advantages: interpretable, verifiable, and actionable. 

\textbf{Task definition: }
Given datasets $\{(X_i, Y_i)\}_{i=1}^N$ from $N$ communities (subreddits), where $X_i = \{x_i^1, x_i^2, ..., x_i^{n_i}\}$ represents the set of texts from community $i$ and $Y_i = \{y_i^1, y_i^2, ..., y_i^{n_i}\}$ denotes the corresponding binary labels, our goal is to build a vocabulary $V = \{v_1, v_2, ..., v_{|V|}\}$ and community-specific score matrices $S_i \in \mathbb{R}^{|V|}$ for each community $i$, where each element $s_i^j$ indicates the contribution of term $v_j$ to moderation decisions in community $i$.
% \end{adjustwidth}

\subsection{Dataset}
We use the Reddit moderation data that was used in prior works in norms discovery~\cite{chandrasekharan2018internet}, so that our finding is comparable with them.
The dataset was collected from May 2016 to March 2017. First, they streamed comments via Reddit's API, then after a 24-hour delay checked which ones had been removed by moderators, and retrieved the original content from their logs. Through this process, about 2.8 million moderated comments from 100 top subreddits. As this dataset only contains moderated comments, we augmented it with comments collected from dumps of 2016 Oct to Nov. 
We built a balanced dataset across 97 subreddits, where the training data ranges from 2,600 to 248,000 instances with a median of 17,928 (\autoref{tab:subreddit_distribution}).

\subsection{Method overview}
The traditional way of building a machine learning model based on n-gram features suffers from the curse of dimensionality and data sparsity. Thus, we propose to use the novel approach PAT to train a neural classifier to extract n-gram features. We employ Partial Attention Transformer (PAT)~\cite{condnli}, which is a model designed for model explanations for text-pair classification tasks, such as natural language inference, and query-document relevance~\cite{kim2024discovering}.

The key strength of PAT lies in that it is trained with labels for full texts, while it is forced to predict the label based on scores from two parts of the texts, which gives it the ability to assign well-calibrated (0-1 range) probability values. After PAT is trained, it is applied at the vocabulary level, enumerating possible term candidates and predicting scores on each term to find out which terms are highly indicative of moderation outcomes.

\subsection{PAT training}
\begin{figure}[htb]
    \centering
    \resizebox{0.8\columnwidth}{!}{
    \begin{tikzpicture}[node distance=0mm]
    \tikzset{Token/.style={rectangle, fill=blue!10,minimum height=4mm, font=\small}}
    \tikzset{PToken/.style={rectangle, fill=blue!10,minimum height=4mm, minimum width=12    mm, font=\small}}
    \tikzset{Predictor/.style={rectangle,draw,minimum width=2.0cm,minimum height=1cm}}
    \tikzset{Prediction/.style={rectangle, fill=black!10, minimum height=3mm, minimum 
    width=3mm, font=\small}}
        
    \matrix[anchor=west, row sep=1mm] (left_tower) {
        \node[Prediction] (z1){$\rho_1$}; \\ [1mm]
        \node[Predictor] (ls1) {BERT}; \\
        \node(t1)[PToken, minimum width=15mm] {$t_1$}; 
        \\
    };
    
    \node[above=15mm of t1] (zs) {};
    \draw[<-] (z1.south) -| ++(0, -0.20);

    \node[right=of left_tower] (center) {};
    \matrix[anchor=west, row sep=1mm, right=of center] {
        \node[Prediction] (z2){$\rho_2$}; \\ [1mm]
        \node[Predictor] (ls2) {BERT}; \\
        \node(t2)[PToken, minimum width=15mm] {$t_2$}; 
        \\
    };

    \node[font=\footnotesize, rectangle, fill=blue!10,  below=20mm of z1] (text_1_head) {$t_1$};	
    \node[font=\footnotesize, rectangle, fill=gray!10, minimum width=18mm, right=of text_1_head] (text_1_body) {You asked a [MASK] question.};

    \node[font=\footnotesize, rectangle, fill=blue!10, below=1mm of text_1_head] (text_2_head) {$t_2$};
    \node[font=\footnotesize, rectangle, fill=gray!10, right=of text_2_head] (htext_2) {stupid};

    \node[rectangle,draw, right=7mm of z1] (combine) {\small{$Agg$}};
    \node[Prediction, above=2mm of combine] (y) {$y$};
    \draw[<-] (z2.south) -| ++(0, -0.20);

    \draw[->] (z1) -- (combine);
    \draw[->] (z2) -- (combine);
    \draw[->] (combine) -- (y);
 
    \end{tikzpicture}
    }
    \caption{PAT training architecture. A comment text ``You asked a stupid question'' is partitioned into two sequences and encoded by BERT. The model is supervised with final $y$ label, while encouraging the model to generate corresponding scores $\rho_1$ and $\rho_2$ for each sequences.  } 
    \label{fig:pat}
\end{figure}
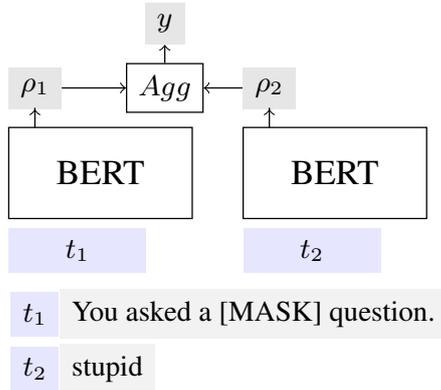

Let $(t, y)$ be a data point in the training dataset, where $t$ is the input text and $y$ is its label. To train PAT, we partition $t$ by randomly selecting two indices $i_s$ and $i_e$, where $i_s$ ≤ $i_e$. This creates two sequences: $t_1$ consist of tokens from position $i_s$ to $i_e$, while $t_2$ is formed by concatenating the remaining portion of $t$ with a [MASK] token in between: token 1 to $i_s$ − 1, followed by [Mask], followed by token $i_e$ + 1 to the end.

PAT encode each partial sequence through a BERT encoder~\cite{devlin2019bert}. The CLS token representation is projected to obtain scores for both classes
\begin{equation}
\text{PAT}(t_i) = W \cdot \text{BERT}_{CLS}(t_i) + b .
\end{equation}

Two outputs from PAT for each of $t_1$ and $t_2$ are aggregated by element-wise sum followed by softmax:

\begin{equation}
    \hat{y} = \text{softmax}(\text{PAT}(t_1) + \text{PAT}(t_2))~\label{eq:logitsum}
\end{equation}
We use cross-entropy loss on predicted probabilities $\hat{y}$ and gold label $y$. 

Given an arbitrary span $t$, we can use PAT predict a probability that a text is moderated if it contain $t$, which is given as 
\begin{equation}
    P(y|t) = \text{softmax}(\text{PAT}(t)) . ~\label{eq:single_softmax}
\end{equation}

\subsection{PAT Inference}
\label{sec:span_clf}
During training, we used the combined score from two sequences $y$ as the output of the model (\autoref{eq:logitsum}). In contrast, when we predict a score for a term, the term alone is used as the only input sequence $t$, and the corresponding output $\rho$ is used as a score for the term after softmax is applied (\autoref{eq:single_softmax}). In this case, $t_2$ does not exist and $\rho_2$ is not computed. 

To evaluating performance of PAT in text classification accuracy, we consider two options. 
The first option, namely PAT (Bipartite), follows the same architecture as in training. It divides the input text into two parts and using the combined score $y$ as the prediction for the text.

However, we are interested in using a term-level score $\rho$. There is a risk that the score $\rho_1$ is not accurate when $t_1$ is a very short sequence. This may not be notable in  PAT (Bipartite), as the other text $t_2$ will be longer when $t_1$ is short, and its output $\rho_2$ can be the dominant factor in deciding $y$.

Thus, we design a second option, namely PAT (Window), which better represents a predictive power with shorter sequences. Given a text, we tokenize it by whitespace and enumerate all three-token windows. Each three-token span is scored by PAT, and these scores are averaged to produce a final prediction. This approach allows us to quantify how much predictive performance is maintained when the model is constrained to shorter spans than those used during training.

\subsection{\mytable construction}
\label{sec:span_extraction}
% -- Begin AI text
For each subreddit, we train one PAT model with the training data from the subreddit. We then build a vocabulary that is shared across different subreddits, which allow us to compare moderation criteria between communities.

Since the potential space of all possible token sequences is prohibitively large, we selectively identify the spans that likely influence moderation decisions.
For each subreddit $i$, we sampled 1,000 comments and applied the corresponding model, $\text{PAT}_{i}$ to these comments. We then extracted spans of texts that received high scores. For each span, we tokenized it and collected n-grams (sequences of $n$ consecutive tokens, where $n$ ranges from 1 to 9) that are substrings of it.

We applied this process across 60 subreddits and built a candidate vocabulary. For each $n$ values, we selected the top 10,000 most frequent n-gram terms, based on probability scoring from an off-the-shelf large language model, Llama-3~\cite{grattafiori2024llama}. 

Finally, we apply each $PAT_{i}$ to score all terms in vocabulary, to get a score matrix $M$, where $M_{i,j}$ indicates the score that PAT for subreddit $i$ has assigned to a term $j$. We refer to $M$ as \mytable throughout the paper. 
\autoref{tab:samples} shows a selected example entries that demonstrate significant differences between subreddits. 
\begin{table}
\centering
\small
\begin{tabular}{l|c|ccc}
\toprule
             & \textbf{mean} & \textbf{politics} & \textbf{Games} & \textbf{history} \\
\midrule
Russian                     & 0.56 & 0.96     & 0.73  & 0.69    \\
\footnotesize{do you speak english}        & 0.59 & 0.94     & 0.88  & 0.55    \\
Tesla                       & 0.49 & 0.36     & 0.84  & 0.82    \\
Asian man                   & 0.68 & 0.49     & 0.96  & 0.76    \\
fucked up                   & 0.65 & 0.51     & 0.81  & 0.93    \\
holy shit                   & 0.50 & 0.51     & 0.78  & 0.84    \\
Trump                       & 0.79 & 0.58     & 1.00  & 0.93   \\
\bottomrule
\end{tabular}
\caption{Sample entries from \mytable . The term ``Russian'' and ``do you speak english'' have high scores for being moderated in the politics subreddit, as users used them to accuse others as a Russian spy. The comments related to Trump were moderated in most subreddits at 2016~\cite{chandrasekharan2018internet}.  }
\label{tab:samples}
\end{table}

\section{Evaluation}
\label{sec:exp}
In this section, we compare the classification performance of various models, including PAT, to evaluate their effectiveness and gain insights about the data. To accept the output of PAT as moderation criteria, we want to make sure that PAT is predictive of moderation outcomes.

We consider two inference methods for PAT that we described in \autoref{sec:span_clf}.
The first method PAT (Bipartite) divides an input text into two parts, making either one or both of parts to be a relatively long sequence. In contrast, PAT (Window) generate multiples sequences as n-gram window. 
Note that both methods actually use the same model parameters. PAT (Window) is representative of the accuracy of short terms in \mytable, while PAT (Bipartite) shows upper bounds for much longer terms.

We evaluated performance across many other models with varying degrees of subreddit awareness:

\textbf{Subreddit aware models:} Models that have knowledge of the specific subreddit being classified, including BERT fine-tuned (FT) on each target subreddit's training data, PAT (Bipartite) and standard PAT trained on subreddit-specific data, ChatGPT prompted with the subreddit name, ChatGPT prompted with both subreddit name and official rules collected via API, LlamaGuard2~\cite{inan2023llama} with subreddit name in the input, and LlamaGuard2 (Toxic) with toxicity definitions and subreddit name in the prompt.

\textbf{Subreddit agnostic models:} Models trained or prompted without information about which specific subreddit the content belongs to, including BERT (FT) trained on aggregated data from 60 subreddits, PAT trained on aggregated data without subreddit identifiers, ChatGPT without subreddit-specific context, LlamaGuard2 with default configuration, and LlamaGuard2 (Toxic) with only toxicity definitions in the prompt.

All models were evaluated across 97 subreddits\footnote{We excluded three subreddits that no longer exist: Incels, soccerstreams, and The\_Donald.}, with 100 instances per subreddit. F1 and AUC scores were computed for each subreddit and then averaged across all communities.

To compute AUC for generative models, we used the token probability for the answer token-the first token that differentiate the classification outcomes-as the prediction score, following the approach used in LlamaGuard~\cite{inan2023llama}. 

\begin{table}[]
\centering
\begin{tabular}{l|cc}
\toprule
                    & \textbf{F1} & \textbf{AUC} \\ \hline
\multicolumn{3}{c}{\textbf{Subreddit aware}}     \\ \hline
BERT (FT)           & 0.81        & 0.90         \\
PAT (Bipratite)     & 0.80        & 0.89         \\
PAT (Window)        & 0.69        & 0.83         \\
ChatGPT             & 0.70        & 0.72         \\
ChatGPT + Rule      & 0.68        & 0.63         \\
LlamaGuard2         & 0.18        & 0.41         \\
LlamaGuard2 (Toxic) & 0.27        & 0.35         \\ \hline
\multicolumn{3}{c}{\textbf{Subreddit agnostic}}  \\ \hline
BERT (FT)           & 0.73        & 0.80         \\
PAT (Window)        & 0.70        & 0.74         \\
ChatGPT             & 0.56        & 0.59         \\
LlamaGuard2         & 0.17        & 0.42         \\
LlamaGuard2 (Toxic) & 0.26        & 0.35        \\
\bottomrule
\end{tabular}
\caption{Comparison of classification performance between BERT and PAT using F1 and AUC metrics.}
\label{tab:f1auc}
\end{table}

\autoref{tab:f1auc} demonstrates that while PAT does not achieve the same performance level as BERT (which has a full text view), it still attains reasonable scores comparable to other methods. Its F1 and AUC metrics are on par with ChatGPT despite the latter having an order of magnitude more parameters.

Notably, PAT (Bipartite) maintains nearly identical performance to BERT (FT), demonstrating the robustness of the PAT training approach.
These results confirm the effectiveness of PAT on Reddit moderation data, which is consistent with its strong performance across diverse tasks in previous work~\cite{condnli, kim2024discovering}. The architecture's success across multiple datasets and tasks indicates that our findings are not limited to this specific Reddit dataset.

Across all models, subreddit-aware variants consistently outperformed their subreddit-agnostic counterparts.

Interestingly, ChatGPT with access to explicit subreddit rules did not outperform the version that only knew the subreddit names. This can be attributed to two factors: First, the majority (70\%) of violations~\cite{park2021detecting} are widely recognized problematic behaviors such as incivility, hate speech, or spam, which ChatGPT already identifies as inappropriate. Second, the actual moderation criteria for more specific rules often cannot be inferred from the officially stated rules alone.

LlamaGuard 2 models performed poorly across all configurations, indicating a significant mismatch between Reddit and their original purpose of safeguarding LLM inputs and outputs.

\section{\mytable\ Analysis}

\mytable constructed using PAT scores provides insights into content moderation patterns across different subreddits.
In this section, we address three critical questions about these patterns: 1) What patterns are used for prediction when an official rule is not sufficient for deciding moderation. 2) Which norms exhibit unexpected tolerance levels across different communities? 3) What meaningful subcategories exist within broadly defined norms?

\subsection{Subreddit specific norms}

\begin{table}[]
\small
\begin{tabular}{c|ccc}
\toprule
\textbf{Subreddit} & \textbf{BERT (FT)} & \textbf{\begin{tabular}[c]{@{}c@{}}ChatGPT \\ (+Subreddit) \end{tabular}} & \textbf{\begin{tabular}[c]{@{}c@{}}ChatGPT \\ (+Rule)\end{tabular}} \\ \hline
churning           & 0.97               & 0.83                                                                          & 0.85                                                                     \\
conspiracy         & 0.90               & 0.72                                                                          & 0.80                                                                     \\
DIY                & 0.83               & 0.71                                                                          & 0.73                                                                     \\
fantasyfootball    & 0.82               & 0.32                                                                          & 0.48                                                                     \\
Games              & 0.85               & 0.57                                                                          & 0.7                                                      \\ \bottomrule
\end{tabular}
\caption{Classification F1 scores on selected subreddits for BERT (FT), ChatGPT run which is provided with subreddit name, and provided with rules of subreddit }
\label{tab:perf_sb}
\end{table}

\autoref{tab:perf_sb} presents F1 scores for selected subreddits, highlighting cases where in-domain supervised models like BERT (FT) significantly outperform both the standard ChatGPT and the rule-enhanced ChatGPT variant. Although providing official rules improves ChatGPT's performance in some communities, substantial performance gaps remain in many subreddits.

These performance disparities suggest that BERT (FT) and PAT can extract subreddit-specific patterns from moderation data, which likely correspond to distinct community norms. To investigate these patterns further, we analyze \mytable for these subreddits.

We targeted r/fantasyfootball and r/Games as they have largest F1 performance gap between GPT and BERT (FT), which suggests these communities have unique moderation standards.

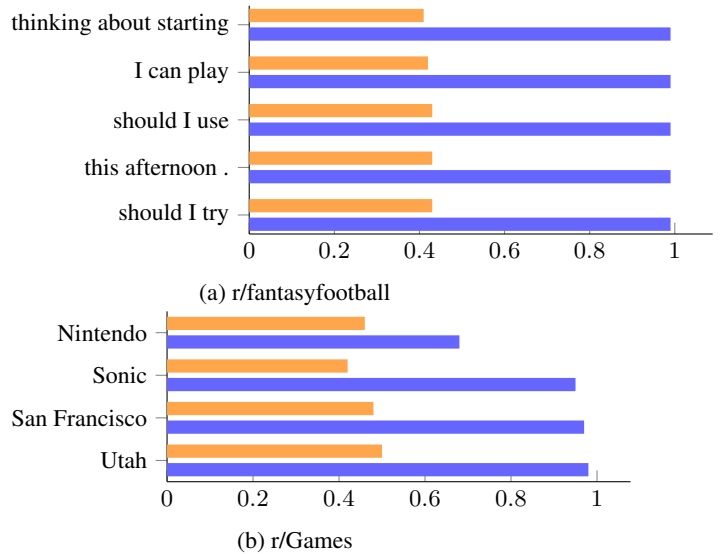
\begin{figure}
    \centering
    \small
    
    % First subfigure - r/fantasyfootball
    \begin{subfigure}[b]{0.48\textwidth}
        \centering
        \begin{tikzpicture}
        \begin{axis}[
        xbar,
        bar width=5pt,
        xmin=0,
        axis lines*=left,
        ytick=data,
        yticklabel style={
            font=\small,
        },
        yticklabels={
        should I try,
        this afternoon .,
        should I use,
        I can play,
        thinking about starting,
        },
        legend style={
            anchor=west,
            draw=none,
            fill=none,
            legend pos=north east,
            legend image post style={scale=0.7},
        },
        width=\textwidth,
        height=0.6\textwidth,
        point meta=rawx
        ]
        % --- Series 1 ---
        \addplot+[draw=none,fill=blue!60] coordinates {
          (0.99,0)
          (0.99,1)
          (0.99,2)
          (0.99,3)
          (0.99,4)
        };
        
        % --- Series 2 ---
        \addplot+[draw=none,fill=orange!70] coordinates {
          (0.43,0)
          (0.43,1)
          (0.43,2)
          (0.42,3)
          (0.41,4)
        };
        \end{axis}
        \end{tikzpicture}
        \caption{r/fantasyfootball}
        \label{fig:fantasyfootball}
    \end{subfigure}
    \hfill
    % second subfigure - r/Games
    \begin{subfigure}[b]{0.48\textwidth}
        \centering
        \begin{tikzpicture}
        \begin{axis}[
        xbar,
        bar width=5pt,
        yshift=2pt,
        xmin=0,
        ymin=-0.5,
        ymax=3.5,
        axis lines*=left,
        ytick=data,
        yticklabel style={
            font=\small,
        },
        yticklabels={
        Utah,
        San Francisco,
        Sonic,
        Nintendo,
        },
        width=\textwidth,
        height=0.5\textwidth,
        point meta=rawx
        ]
        % --- Series 1 ---
        \addplot+[draw=none,fill=blue!60] coordinates {
          (0.98,0)
          (0.97,1)
          (0.95,2)
          (0.68,3)
        };
        
        % --- Series 2 ---
        \addplot+[draw=none,fill=orange!70] coordinates {
          (0.50,0)
          (0.48,1)
          (0.42,2)
          (0.46,3)
        };
        \end{axis}
        \end{tikzpicture}
        \caption{r/Games}
        \label{fig:games}
    \end{subfigure}
    \caption{Representative terms with large score differences between subreddit-specific scores (blue) and average across subreddits (orange) for r/fantasyfootball and r/Games.}
    \label{fig:subreddit_to_mean}
\end{figure}

\autoref{fig:subreddit_to_mean} shows the terms with large score differences between the specific subreddits and the mean across other subreddits. 
For r/fantasyfootball, phrases such as ``thinking about starting'' and ``should I use'' appeared as particularly predictive of moderation decisions in this community.
Examining the official rules of r/fantasyfootball, we found a rule that says ``No individual threads of any kind specific to your team or league.'' 
We can infer that these high-scoring phrases typically appear when users ask questions specific to their own fantasy teams.
Since our dataset lacks features to determine whether these comments appeared in individual threads or within valid threads, we cannot directly verify rule violations.

For r/Games, we found two interesting patterns. First, some game titles and company names have higher scores to be moderated. Manual inspection of comments containing these terms revealed they were often removed for expressing criticism toward game companies.  We expect that moderators had decided to delete these comments as they concluded them not providing additional values to the discussion. We consider this as unique subreddit specific criteria, as such decisions is not obvious predictable from most relevant generic norms ``Low-quality comments'' or ``Be civil''.

Second, location names like ``San Francisco'' or ``Utah'' received unusually high moderation scores in r/Games compared to other subreddits. This pattern reflects the community's particularly strict enforcement of off-topic content rules. While ``No off-topic posts'' is a common norm, these scores indicate particular lower tolerance in r/Games.

Note that high scores for these terms do not imply that moderators will blindly remove comments simply for containing them. Rather, these patterns reveal the underlying criteria (more strict in some topics) why certain comments were moderated in these communities. The ability of supervised models to correctly classify these comments demonstrates that they have learned to recognize the implicit criteria that guide community-specific moderation decisions, even when these criteria are not explicitly stated in official rules.

\subsection{Blaming Moderators}
Criticizing moderators on Reddit is known to frequently result in content removal across most subreddits~\cite{chandrasekharan2018internet, fiesler2018reddit}.
We investigated whether there are differences in how communities determine which moderator-related comments warrant moderation.
 \mytable showed that unigram terms like  ``moderator'' or ``mod'' have high scores at many subreddits. Specifically the term ``mod'' had scores ranging from 0.18 to 0.99, with mean of 0.68, and in 12, ``mod'' received scores over 0.9.  

Based on these observations, we hypothesized that moderation classifiers for many subreddits might be sensitive to any mention of moderators, potentially regardless of context or intent.
To test this hypothesis,
we constructed a dataset of 50 synthetic comments generated by ChatGPT and Claude that contained the keyword ``mod'' without expressing blame or criticism toward moderators.

Applying our trained classifiers to this synthetic dataset, we found that 16 out of 60 classifiers (26.7\%) predicted \textbf{all} 50 instances as requiring moderation, despite their non-critical nature.
This suggests several possibilities: (1) people expressed sarcastic descriptions about moderation, so that any mention about moderation is considered sarcastically blaming moderator. (2) synthetic sentences are not natural, which actually triggered moderation.

To confirm this finding with real-world data, we subsequently analyzed the actual moderation rates of comments containing the term ``mod'' across our dataset. The results strongly supported our hypothesis: in 16\% of subreddits, more than 90\% of comments mentioning ``mod'' were removed, and in 34\% of subreddits, the removal rate exceeded 80\%. These high moderation rates for comments containing a simple reference to moderators support the hypothesis that many communities may have low tolerance for moderator discussions of any kind, potentially explaining why our classifiers flagged even neutral mentions.

\subsection{Tolerance to personal attack}

We hypothesized that there would be meaningful difference in tolerance to a personal attack. 
Personal attack is very broad categories and there are numerous lexicons.
To avoid inspecting all vocabulary, we implemented a clustering approach based on score similarity across subreddits.

We consider that two terms similar in their contribution to moderation decision if their scores over subreddits are highly correlated. 
Using Pearson correlation coefficients as our similarity metric we performed k-means clustering (k=100) on the term space. We additionally filtered terms that are far from the corresponding centroid to ensure purity of clusters. The resulting Silhouette score of 0.15 indicates a weak but present clustering structure.

We manually categorized these clusters according to the following criteria: Personal attacks included terms potentially offensive toward the second person (``you'') or containing general slurs; hate speech includes terms offensive to demographic groups; topical clusters contained topically coherent terms not inherently offensive; URL and markdown clusters focused on structural elements; and the remainder were classified as "other."

\begin{table}[]
\centering
\small
\begin{tabular}{lc}
\toprule
                         & \textbf{\# of clusters} \\ \midrule
\textbf{Personal attack} & 10                      \\
\textbf{Hate speech}     & 6                       \\
\textbf{Topical}         & 16                      \\
\textbf{URL}             & 18                      \\
\textbf{Markdown}        & 5                       \\
\textbf{Others}          & 45                      \\ \bottomrule
\end{tabular}
\caption{Distribution of term clusters categorized by content moderation types, derived from k-means clustering (k=100). }
\label{tab:clusters}
\end{table}

We focused on clusters classified as personal attacks. 
To differentiate between these clusters, we assigned a name to each based on what is common among the terms. We generated descriptive name using Claude~\footnote{https://claude.ai/}.
Note that these names are intended to make distinguishing clusters easy and do not precisely describe the clusters.

\autoref{tab:personal_attack} shows the these personal attack clusters with their average scores over subreddits and standard deviation.
The clustering has successfully differentiated between distinct personal attack types ranging from the most offensive ones like ``direct intelligence insults'' (scoring highest at 0.84) to more subtle forms like epistemic competence undermining and boundary-crossing advice (scoring 0.56-0.57).

\begin{figure}
    \centering
    
\begin{tikzpicture}
\begin{axis}[
    width=\linewidth,
    xlabel={Term clusters ID},
    ylabel={Moderation scores},
    minor grid style={gray!25},
    major grid style={gray!50},
    xmin=0.5, xmax=9.5,
    xtick={1,2,3,4,5,6,7,8,9},
    xticklabels={81, 20, 42, 46, 14, 5, 37, 52, 39},
    ymin=0.3, ymax=1.05,
    ytick={0.3,0.4,0.5,0.6,0.7,0.8,0.9,1.0},
    mark size=2pt,
    line width=1pt
]

% Series 1
\addplot[blue, mark=*] coordinates {
    (1, 0.99) (2, 0.98) (3, 0.98) (4, 0.96) (5, 0.97) (6, 0.91) (7, 0.6) (8, 0.86) (9, 0.77)
};

% Series 2
\addplot[red, mark=square*] coordinates {
    (1, 0.98) (2, 0.98) (3, 0.96) (4, 0.91) (5, 0.95) (6, 0.93) (7, 0.9) (8, 0.57) (9, 0.82)
};

% Series 3
\addplot[green!60!black, mark=triangle*] coordinates {
    (1, 0.91) (2, 0.9) (3, 0.89) (4, 0.86) (5, 0.86) (6, 0.82) (7, 0.76) (8, 0.73) (9, 0.75)
};

% Series 4
\addplot[orange, mark=diamond*] coordinates {
    (1, 0.77) (2, 0.76) (3, 0.73) (4, 0.71) (5, 0.74) (6, 0.72) (7, 0.7) (8, 0.72) (9, 0.7)
};

% Series 5
\addplot[purple, mark=pentagon*] coordinates {
    (1, 0.94) (2, 0.86) (3, 0.89) (4, 0.87) (5, 0.74) (6, 0.56) (7, 0.49) (8, 0.53) (9, 0.55)
};

% Series 6
\addplot[brown, mark=oplus*] coordinates {
    (1, 0.95) (2, 0.92) (3, 0.88) (4, 0.68) (5, 0.78) (6, 0.73) (7, 0.57) (8, 0.51) (9, 0.49)
};

\end{axis}
\end{tikzpicture}
\caption{Average moderation scores for personal attack term clusters across six different subreddits. Each line represents a distinct subreddit (anonymized), showing how communities vary in their tolerance for different types of personal attacks.  }
\label{fig:pa_sb}
\end{figure}
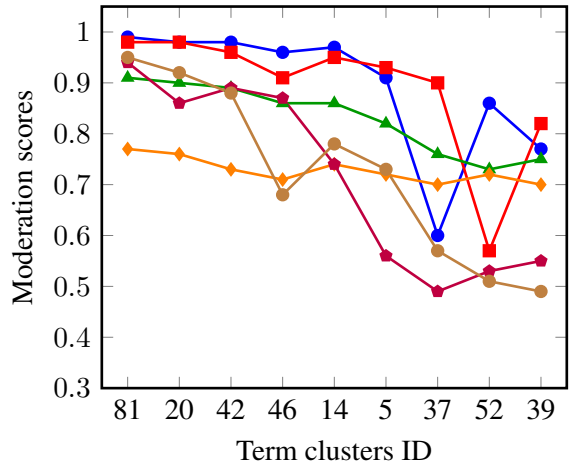

\autoref{fig:pa_sb} shows how moderation scores for each personal attack cluster vary across six different subreddits. While the overall pattern shows that some subreddits consistently maintain higher or lower moderation thresholds across all clusters, we observe exceptions in some subreddits.
Two subreddits show much lower scores for clusters 37 and 52, respectively. This pattern suggests community-specific tolerance for certain types of personal address. For instance, the lower scores for cluster 52 (boundary-crossing advice) likely indicate that in advice-focused communities, direct guidance that might be considered intrusive elsewhere is instead viewed as appropriate. Similarly, the reduced sensitivity to cluster 37 (second-person framing) in another subreddit suggests that direct addressing of other users is more acceptable within its conversational norms.

\begin{table}[h!]
\centering
\small
\begin{tabular}{@{}c c l@{}}
\toprule
\textbf{ID} & \textbf{Score (SD)} & \textbf{Name} \\ \midrule
81  & 0.84 (0.15) & Direct intelligence insults \\
20  & 0.81 (0.16) & Profane command attacks \\
42  & 0.78 (0.16) & Behavioral mockery \\
46  & 0.72 (0.16) & Indirect intelligence attacks \\
14  & 0.71 (0.16) & Dismissive commands \\
5   & 0.64 (0.15) & Identity questioning \\
37  & 0.57 (0.13) & Second-person framing \\
52  & 0.57 (0.13) & Boundary-crossing advice \\
39  & 0.56 (0.14) & Competence undermining \\
\bottomrule
\end{tabular}
\caption{Clusters of personal-attack (toxic) language with average scores and standard deviations. Terms for each clusters are listed in \autoref{tab:pa_full}.}
\label{tab:personal_attack}
\end{table}

Our analysis of gives a revealed several important patterns in content moderation. 

\subsection{Summary}
Our analysis of \mytable reveals three key insights about content moderation across Reddit communities.

First, for community-specific rules, we found patterns that strongly predict moderation decisions in our dataset but may not represent objectively sufficient conditions for removal. This suggests supervised models may learn spurious correlations rather than moderators' true intent.

Second, our ``blaming moderators'' analysis demonstrated how historical moderation data can lead classifiers to flag even
neutral mentions of moderators as violations.
This pattern might not align with moderators' future expectations, highlighting needs for potential disconnects between learned patterns and intended policy.

Finally, our clustering of personal attacks revealed that toxic content exists on a spectrum, with communities showing varying tolerance levels for different types of attacks. While prior work often treats toxicity as having universal criteria and thresholds, our findings suggest that community-specific calibration of tolerance for different attack types could improve moderation effectiveness.

\section{Conclusion}

In this paper, we introduced a novel approach to understanding content moderation criteria across different online communities by leveraging an interpretable architecture PAT, to extract lexical expressions predictive of moderation decisions. 
These expressions provide insights into classification criteria while functioning effectively as classifiers themselves.
Our methodology could benefit other classification tasks, which has multiple sub-domains with possible different classifications criteria by assisting developers to understand the underlying criteria and discover possible biases of the dataset.

\section{Acknowledgements}
This work was supported by the McIntire School of Commerce Foundation and by a Faculty AI Research Award from the University of Virginia's Darden School of Business.
We thank the University of Virginia's High Performance Computing team for providing excellent resources for conducting our experiments.

\section*{Limitations}
Our analysis and dataset have several limitations that could affect the generalizability of our findings.

First, our approach only captures patterns observable in short textual expressions, excluding toxic behaviors that require broader context to identify. Our data also lacks important contextual elements such as previous comments or parent posts that often influence moderation decisions.

Second, our use of balanced datasets with equal proportions of moderated and non-moderated content differs significantly from the natural distribution on Reddit, where only about 5\% of comments are typically moderated. This sampling approach, while standard for classification tasks, may amplify certain patterns that would be less prominent in real-world applications.

Third, the dataset's limited time window and size may be critical to the resulting models having suboptimal performance and biases. Some patterns identified in our analysis might not persist or might appear different with more comprehensive data collection spanning longer periods.

These limitations suggest caution in interpreting our findings and highlight opportunities for future work with richer contextual data, more representative sampling, and longitudinal analysis of moderation patterns.

% Bibliography entries for the entire Anthology, followed by custom entries
%\bibliography{anthology,custom}
% Custom bibliography entries only
\bibliography{custom}

\begin{thebibliography}{23}
\providecommand{\natexlab}[1]{#1}

\bibitem[{Ahmed et~al.(2024)Ahmed, Kaiser, Hossain, and Andersson}]{ahmed2024comparative}
Shamim Ahmed, M~Shamim Kaiser, Mohammad~Shahadat Hossain, and Karl Andersson. 2024.
\newblock A comparative analysis of lime and shap interpreters with explainable ml-based diabetes predictions.
\newblock \emph{IEEE Access}.

\bibitem[{Burkart and Huber(2021)}]{burkart2021survey}
Nadia Burkart and Marco~F Huber. 2021.
\newblock A survey on the explainability of supervised machine learning.
\newblock \emph{Journal of Artificial Intelligence Research}, 70:245--317.

\bibitem[{Chandrasekharan et~al.(2019)Chandrasekharan, Gandhi, Mustelier, and Gilbert}]{chandrasekharan2019crossmod}
Eshwar Chandrasekharan, Chaitrali Gandhi, Matthew~Wortley Mustelier, and Eric Gilbert. 2019.
\newblock Crossmod: A cross-community learning-based system to assist reddit moderators.
\newblock \emph{Proceedings of the ACM on human-computer interaction}, 3(CSCW):1--30.

\bibitem[{Chandrasekharan et~al.(2018)Chandrasekharan, Samory, Jhaver, Charvat, Bruckman, Lampe, Eisenstein, and Gilbert}]{chandrasekharan2018internet}
Eshwar Chandrasekharan, Mattia Samory, Shagun Jhaver, Hunter Charvat, Amy Bruckman, Cliff Lampe, Jacob Eisenstein, and Eric Gilbert. 2018.
\newblock The internet's hidden rules: An empirical study of reddit norm violations at micro, meso, and macro scales.
\newblock \emph{Proceedings of the ACM on Human-Computer Interaction}, 2(CSCW):1--25.

\bibitem[{Devlin et~al.(2019)Devlin, Chang, Lee, and Toutanova}]{devlin2019bert}
Jacob Devlin, Ming-Wei Chang, Kenton Lee, and Kristina Toutanova. 2019.
\newblock Bert: Pre-training of deep bidirectional transformers for language understanding.
\newblock In \emph{Proceedings of the 2019 conference of the North American chapter of the association for computational linguistics: human language technologies, volume 1 (long and short papers)}, pages 4171--4186.

\bibitem[{Fiesler et~al.(2018)Fiesler, Jiang, McCann, Frye, and Brubaker}]{fiesler2018reddit}
Casey Fiesler, Jialun Jiang, Joshua McCann, Kyle Frye, and Jed Brubaker. 2018.
\newblock Reddit rules! characterizing an ecosystem of governance.
\newblock In \emph{Proceedings of the International AAAI Conference on Web and Social Media}, volume~12.

\bibitem[{Goyal et~al.(2024)Goyal, Lambert, Jain, and Chandrasekharan}]{goyal2024uncovering}
Agam Goyal, Charlotte Lambert, Yoshee Jain, and Eshwar Chandrasekharan. 2024.
\newblock Uncovering the internet's hidden values: An empirical study of desirable behavior using highly-upvoted content on reddit.
\newblock \emph{arXiv preprint arXiv:2410.13036}.

\bibitem[{Grattafiori et~al.(2024)Grattafiori, Dubey, Jauhri, Pandey, Kadian, Al-Dahle, Letman, Mathur, Schelten, Vaughan et~al.}]{grattafiori2024llama}
Aaron Grattafiori, Abhimanyu Dubey, Abhinav Jauhri, Abhinav Pandey, Abhishek Kadian, Ahmad Al-Dahle, Aiesha Letman, Akhil Mathur, Alan Schelten, Alex Vaughan, et~al. 2024.
\newblock The llama 3 herd of models.
\newblock \emph{arXiv preprint arXiv:2407.21783}.

\bibitem[{Hartvigsen et~al.(2022)Hartvigsen, Gabriel, Palangi, Sap, Ray, and Kamar}]{hartvigsen2022toxigen}
Thomas Hartvigsen, Saadia Gabriel, Hamid Palangi, Maarten Sap, Dipankar Ray, and Ece Kamar. 2022.
\newblock Toxigen: A large-scale machine-generated dataset for adversarial and implicit hate speech detection.
\newblock \emph{arXiv preprint arXiv:2203.09509}.

\bibitem[{Inan et~al.(2023)Inan, Upasani, Chi, Rungta, Iyer, Mao, Tontchev, Hu, Fuller, Testuggine et~al.}]{inan2023llama}
Hakan Inan, Kartikeya Upasani, Jianfeng Chi, Rashi Rungta, Krithika Iyer, Yuning Mao, Michael Tontchev, Qing Hu, Brian Fuller, Davide Testuggine, et~al. 2023.
\newblock Llama guard: Llm-based input-output safeguard for human-ai conversations.
\newblock \emph{arXiv preprint arXiv:2312.06674}.

\bibitem[{Kim et~al.(2023)Kim, Rahimi, and Allan}]{condnli}
Youngwoo Kim, Razieh Rahimi, and James Allan. 2023.
\newblock \href {https://aclanthology.org/2023.findings-emnlp.456} {Conditional natural language inference}.
\newblock In \emph{Findings of the Association for Computational Linguistics: EMNLP 2023}, pages 6833--6851, Singapore. Association for Computational Linguistics.

\bibitem[{Kim et~al.(2024)Kim, Rahimi, and Allan}]{kim2024discovering}
Youngwoo Kim, Razieh Rahimi, and James Allan. 2024.
\newblock Discovering biases in information retrieval models using relevance thesaurus as global explanation.
\newblock In \emph{Proceedings of the 2024 Conference on Empirical Methods in Natural Language Processing}, pages 19530--19547.

\bibitem[{Kolla et~al.(2024)Kolla, Salunkhe, Chandrasekharan, and Saha}]{kolla2024llm}
Mahi Kolla, Siddharth Salunkhe, Eshwar Chandrasekharan, and Koustuv Saha. 2024.
\newblock Llm-mod: Can large language models assist content moderation?
\newblock In \emph{Extended Abstracts of the CHI Conference on Human Factors in Computing Systems}, pages 1--8.

\bibitem[{Lambert et~al.(2024)Lambert, Choi, and Chandrasekharan}]{lambert2024positive}
Charlotte Lambert, Frederick Choi, and Eshwar Chandrasekharan. 2024.
\newblock " positive reinforcement helps breed positive behavior": Moderator perspectives on encouraging desirable behavior.
\newblock \emph{Proceedings of the ACM on Human-Computer Interaction}, 8(CSCW2):1--33.

\bibitem[{Lundberg and Lee(2017)}]{lundberg2017unified}
Scott~M Lundberg and Su-In Lee. 2017.
\newblock A unified approach to interpreting model predictions.
\newblock \emph{Advances in neural information processing systems}, 30.

\bibitem[{Neuman and Cohen(2023)}]{neuman2023ai}
Yair Neuman and Yochai Cohen. 2023.
\newblock Ai for identifying social norm violation.
\newblock \emph{Scientific Reports}, 13(1):8103.

\bibitem[{Park et~al.(2024)Park, Li, Jung, Volkova, Mitra, Jurgens, and Tsvetkov}]{park2024valuescope}
Chan~Young Park, Shuyue~Stella Li, Hayoung Jung, Svitlana Volkova, Tanu Mitra, David Jurgens, and Yulia Tsvetkov. 2024.
\newblock Valuescope: Unveiling implicit norms and values via return potential model of social interactions.
\newblock In \emph{Findings of the Association for Computational Linguistics: EMNLP 2024}, pages 16659--16695.

\bibitem[{Park et~al.(2021)Park, Mendelsohn, Radhakrishnan, Jain, Kanakagiri, Jurgens, and Tsvetkov}]{park2021detecting}
Chan~Young Park, Julia Mendelsohn, Karthik Radhakrishnan, Kinjal Jain, Tushar Kanakagiri, David Jurgens, and Yulia Tsvetkov. 2021.
\newblock Detecting community sensitive norm violations in online conversations.
\newblock In \emph{Findings of the Association for Computational Linguistics: EMNLP 2021}, pages 3386--3397.

\bibitem[{Phillips et~al.(2018)Phillips, Chang, and Friedler}]{phillips2018interpretable}
Richard Phillips, Kyu~Hyun Chang, and Sorelle~A Friedler. 2018.
\newblock Interpretable active learning.
\newblock In \emph{Conference on fairness, accountability and transparency}, pages 49--61. PMLR.

\bibitem[{Rao et~al.(2023)Rao, Jiang, Pyatkin, Gu, Tandon, Dziri, Brahman, and Choi}]{rao-etal-2023-makes}
Kavel Rao, Liwei Jiang, Valentina Pyatkin, Yuling Gu, Niket Tandon, Nouha Dziri, Faeze Brahman, and Yejin Choi. 2023.
\newblock \href {https://doi.org/10.18653/v1/2023.findings-emnlp.812} {What makes it ok to set a fire? iterative self-distillation of contexts and rationales for disambiguating defeasible social and moral situations}.
\newblock In \emph{Findings of the Association for Computational Linguistics: EMNLP 2023}, pages 12140--12159, Singapore. Association for Computational Linguistics.

\bibitem[{Ribeiro et~al.(2016)Ribeiro, Singh, and Guestrin}]{lime}
Marco~Tulio Ribeiro, Sameer Singh, and Carlos Guestrin. 2016.
\newblock \href {https://doi.org/10.1145/2939672.2939778} {"why should i trust you?": Explaining the predictions of any classifier}.
\newblock In \emph{Proceedings of the 22nd ACM SIGKDD International Conference on Knowledge Discovery and Data Mining}, KDD '16, page 1135–1144, New York, NY, USA. Association for Computing Machinery.

\bibitem[{R{\"o}ttger et~al.(2021)R{\"o}ttger, Vidgen, Nguyen, Waseem, Margetts, and Pierrehumbert}]{rottger2021hatecheck}
Paul R{\"o}ttger, Bertie Vidgen, Dong Nguyen, Zeerak Waseem, Helen Margetts, and Janet Pierrehumbert. 2021.
\newblock Hatecheck: Functional tests for hate speech detection models.
\newblock In \emph{Proceedings of the 59th Annual Meeting of the Association for Computational Linguistics and the 11th International Joint Conference on Natural Language Processing (Volume 1: Long Papers)}, pages 41--58.

\bibitem[{Weld et~al.(2024)Weld, Zhang, and Althoff}]{weld2024making}
Galen Weld, Amy~X Zhang, and Tim Althoff. 2024.
\newblock Making online communities ‘better’: a taxonomy of community values on reddit.
\newblock In \emph{Proceedings of the International AAAI Conference on Web and Social Media}, volume~18, pages 1611--1633.

\end{thebibliography}
\newpage

\appendix

\section{Dataset details}

\begin{table}[h!]
\centering
\begin{tabular}{lcp{4cm}}
\toprule
\textbf{Range} & \textbf{Count} & \textbf{Examples} \\
\midrule
$\geq$100K & 5 & politics, AskReddit, worldnews \\
50K--99K & 7 & relationships, TwoXChromosomes, gonewild \\
20K--49K & 31 & PoliticalDiscussion, askscience, personalfinance \\
10K--19K & 52 & Android, OutOfTheLoop, atheism \\
2K--10K & 2 & NeutralPolitics, EnoughTrumpSpam \\
\bottomrule
\end{tabular}
\caption{Size distribution of training data across the 97 subreddits in our dataset.}
\label{tab:subreddit_distribution}
\end{table}

\section{Baseline implementations}

\subsection*{ChatGPT - Prompt}

For ChatGPT runs in \autoref{sec:exp}, we concatenated a task specific instruction to a target text to be classified.

\textbf{ChatGPT with rule}

For each subreddit, we construct a prompt by concatenating its moderation rules. We used the both summary and detail parts of rules which is collected by the official API. 

The prompt is formatted as follows:

\begin{quote}
\texttt{[Subreddit moderation rules]} \\
\texttt{The above rule describes prohibited contents. Classify if the following text is prohibited. If prohibited, output 'unsafe' as a first token. If not, output 'safe'.}
\end{quote}

The model is instructed to output the token \texttt{unsafe} if the input violates any subreddit rule and \texttt{safe} otherwise.

\textbf{ChatGPT - subreddit aware}

\begin{quote}
\texttt{If the following text is posted in \{subreddit\} subreddit, will it be moderated (deleted)?} \\
\texttt{Answer Yes or No, as a single token.}
\end{quote}

\textbf{ChatGPT - subreddit agnostic}
\begin{quote}
\texttt{If the following text is posted on Reddit, will it be moderated (deleted)?} \\
\texttt{Answer Yes or No, as a single token.}
\end{quote}

\section{PAT model implementations}

\subsection*{Training Hyperparameters}

The following hyperparameters were used during model fine-tuning:

\begin{itemize}
    \item \textbf{Number of training epochs:} 3
    \item \textbf{Learning rate:} 5e-5
    \item \textbf{Training batch size:} 16
    \item \textbf{Warmup ratio:} 0.1
    \item \textbf{Weight decay:} 0.01
    \item \textbf{Learning rate scheduler:} Linear with warmup
\end{itemize}
No extensive hyperparameter tuning was used throughout the development process.

\begin{table*}[h!]
\small
\centering
\begin{tabular}{rrll}
\toprule
\multicolumn{1}{l}{ID} & \multicolumn{1}{l}{Score} & Name                             & Sample Terms                                                                                                                                                                                                                                                                                                 \\ \hline
81                     & 0.84                      & Direct intelligence insults      & \begin{tabular}[c]{@{}l@{}}pussy; whore; are you gay; you are an idiot; you are a moron; \\ trump is a racist; you are a loser; are you a moron; are you gay?; \\ you are a stupid; you are so stupid; what a fucking idiot; \\ are you an idiot; stop being a bitch; stop being an idiot;\end{tabular}      \\ \hline
20                     & 0.81                      & Profane command attacks          & \begin{tabular}[c]{@{}l@{}}motherfucker; motherfuckers; shut the fuck; shut up you; \\ kiss my ass; fuck you,; shut the fuck up; get the fuck out; \\ eat shit and die; calm the fuck down; fucked in the ass; \\ blow your brains out; you are a fucking; get off your ass;\end{tabular}                    \\\hline
42                     & 0.78                      & Behavioral mockery               & \begin{tabular}[c]{@{}l@{}}douchebag; dickhead; slut; idiot; cunt; douchebags; \\ slander; bitches; dick,; dumb and dumber; stupid stupid stupid; \\ what a stupid; a piece of shit; you are a fool; \\ dumb, stupid,; are you a nerd; you are a hypocrite;\end{tabular}                                     \\\hline
46                     & 0.72                      & Indirect intelligence attacks    & \begin{tabular}[c]{@{}l@{}}shut up; yourself a; shut up and; shut up,; change your life; \\ live your life; get yourself a; teach your child; shut up.; \\ eat your heart; "shut up; go away,; stop being a; \\ in your mouth; shut the hell; stop being so; sick of your;\end{tabular}                      \\\hline
14                     & 0.71                      & Dismissive commands              & \begin{tabular}[c]{@{}l@{}}slutty; stupidest; stupid,; stupid people; stupidity is; \\ this is stupid; this is stupid and; what is this stupid;\\  too stupid to be; stupid...; there is no stupid; \\ complete and utter bullshit; kind of an idiot;\end{tabular}                                           \\\hline
5                      & 0.64                      & Identity questioning             & \begin{tabular}[c]{@{}l@{}}you a; are you a; you are a; are you an; you are an; \\ you were a; you are one; like you are; you are such; \\ like you were; you is a; you must be a; you are one of; \\ that you are a; looks like you are; because you are a;\end{tabular}                                    \\\hline
37                     & 0.57                      & Second-person framing            & \begin{tabular}[c]{@{}l@{}}you; youll; youd; you are; are you; you don; you must; \\ - you; you were; all you; because you; you just;\\  maybe you; you,; (you; now you; you.; you all;\\  , you; you never\end{tabular}                                                                                     \\\hline
52                     & 0.57                      & Boundary-crossing advice         & \begin{tabular}[c]{@{}l@{}}get your; find your; keep your; things you; let your; put your;\\  your personal; save your; what your; where your; please dont; \\ leaving your; please do not; for all your; speak to a;\\  with all your; let it go; for your personal; part of your; speak to an\end{tabular} \\\hline
39                     & 0.56                      & Competence undermining & \begin{tabular}[c]{@{}l@{}}you may not; you should have; you have no; \\ you dont know; the reason you; you should just; \\ you had no; you dont get; you want to be;\\  you do not need; you have no idea; you are not going;\end{tabular}         \\ \bottomrule                                                        
\end{tabular}
\caption{Clusters of personal-attack (toxic) language with sample terms}
\label{tab:pa_full}

\end{table*}

\section{Disclaimers}

\subsection{Reddit Content Moderation Dataset}

The dataset of Reddit moderation~\cite{chandrasekharan2018internet}, used as the artifact in this paper, has been carefully curated and anonymized by its creators to protect user privacy and prevent the inclusion of personally identifying information. The dataset consists of comments text with personal identifiable meta data removed. 

\subsection{AI Assistance}
We acknowledge the use of AI assistants, Claude by Anthropic~\footnote{https://www.anthropic.com/claude} and GPT by OpenAI~\footnote{https://chat.openai.com/}, in the writing process of this paper. These AI assistants provided support in drafting and refining the contents of the paper. However, all final decisions regarding the content, structure, and claims were made by the human authors, who carefully reviewed and edited the generated content.

\subsection{Computational cost}
We used one of the following GPUs for training: NVIDIA RTX A6000, A40, or V100. With any of these devices, training took less than two hours. Note that all implementations are designed to run within 16 GB of VRAM, and the computational cost is typical compared to the standard practice of fine-tuning the BERT-base-uncased model from Hugging Face's Transformers library.

\end{document}